\documentclass[letterpaper, 10 pt, conference]{ieeeconf}
\IEEEoverridecommandlockouts
\usepackage{times}
\usepackage{helvet}
\usepackage{courier}
\usepackage[hyphens]{url}
\usepackage{graphicx}
\usepackage{caption}
\usepackage{cite}
\usepackage{amsmath}
\usepackage{amsfonts}
\usepackage{booktabs}
\usepackage{multirow}
\usepackage{makecell}
\usepackage{subcaption}
\usepackage{algorithm}
\usepackage{algorithmic}
\usepackage{newfloat}
\usepackage{listings}
\usepackage{color}

\usepackage[pagebackref=false,breaklinks=true,colorlinks,bookmarks=true,bookmarksnumbered=true]{hyperref}
\title{\LARGE \bf
Spike-EVPR: Deep Spiking Residual Networks with SNN-Tailored Representations for Event-Based Visual Place Recognition
}

\author{Zuntao Liu$^{*}$, Yaohui Li$^{*}$, Chenming Hu, Delei Kong, Junjie Jiang, Zheng Fang$^{\dagger}$
\thanks{This work was supported by the National Natural Science Foundation of China under Grants 62073066, the Fundamental Research Funds for Central Universities under Grant N2226001, and 111 Project under Grant B16009. (Corresponding author: Zheng Fang, e-mail: fangzheng@mail.neu.edu.cn)}
\thanks{Zuntao Liu, Yaohui Li, Chenming Hu, Junjie Jiang, and Zheng Fang are with Faculty of Robot Science and Engineering, Northeastern University, Shenyang, China. Delei Kong is with  School of Artificial Intelligence and Robotics, Hunan University, Changsha, China.
}
\thanks{$^{*}$Equal contributions. $^{\dagger}$Corresponding author.}
}

\begin{document}

\maketitle
\thispagestyle{empty}
\pagestyle{empty}

\begin{abstract}
Event cameras are ideal for visual place recognition (VPR) in challenging environments due to their high temporal resolution and high dynamic range.
However, existing methods convert sparse events into dense frame-like representations for Artificial Neural Networks (ANNs), ignoring event sparsity and incurring high computational cost.
Spiking Neural Networks (SNNs) complement event data through discrete spike signals to enable energy-efficient VPR, but their application is hindered by the lack of effective spike-compatible representations and deep architectures capable of learning discriminative global descriptors. 
To address these limitations, we propose Spike-EVPR, a directly trained, end-to-end SNN framework tailored for event-based VPR.
First, we introduce two complementary event representations, MCS-Tensor and TSS-Tensor, designed to reduce temporal redundancy while preserving essential spatio-temporal cues.
Furthermore, we propose a deep spiking residual architecture that effectively aggregates these features to generate robust place descriptors.
Extensive experiments on the Brisbane-Event-VPR and DDD20 datasets demonstrate that Spike-EVPR achieves state-of-the-art performance, improving Recall@1 by 7.61\% and 13.20\%, respectively, while significantly reducing energy consumption.
\end{abstract}

\section{Introduction}
Visual place recognition (VPR) aims to localize a query image by matching it against geo-referenced database images and is an essential component in robotics and computer vision~\cite{xu2020probabilistic,berton2022deep,lu2024cricavpr}. 
While image-based VPR methods have achieved impressive results~\cite{lu2024towards}, they struggle with motion blur and extreme light conditions~\cite{lowry2015visual,schubert2023visual}.
To address these challenges, event cameras have been increasingly adopted in VPR due to their high temporal resolution, high dynamic range, and low latency~\cite{gallego2020event}, leading to event-based VPR (EVPR)~\cite{fischer2020event,kong2022event,fischer2022many,hou2023fe}.

However, most existing EVPR methods rely on conventional Artificial Neural Networks (ANNs) that convert asynchronous events into dense frame-like representations. This preprocessing undermines the intrinsic sparsity and temporal precision of event streams, leading to computational redundancy and suboptimal efficiency.
In contrast, Spiking Neural Networks (SNNs)~\cite{rueckauer2017conversion, han2020deep, deng2021optimal} naturally align with the temporal dynamics of event streams,
offering energy-efficient computation through event-driven spike processing. 
Unlike ANNs that depend on dense floating-point operations, SNNs process information using discrete binary spikes, maintaining high representational capacity while significantly lowering power consumption.
These properties make SNNs a promising and energy-efficient alternative for EVPR.

\begin{figure}[t]
\centering
\includegraphics[width=\columnwidth]{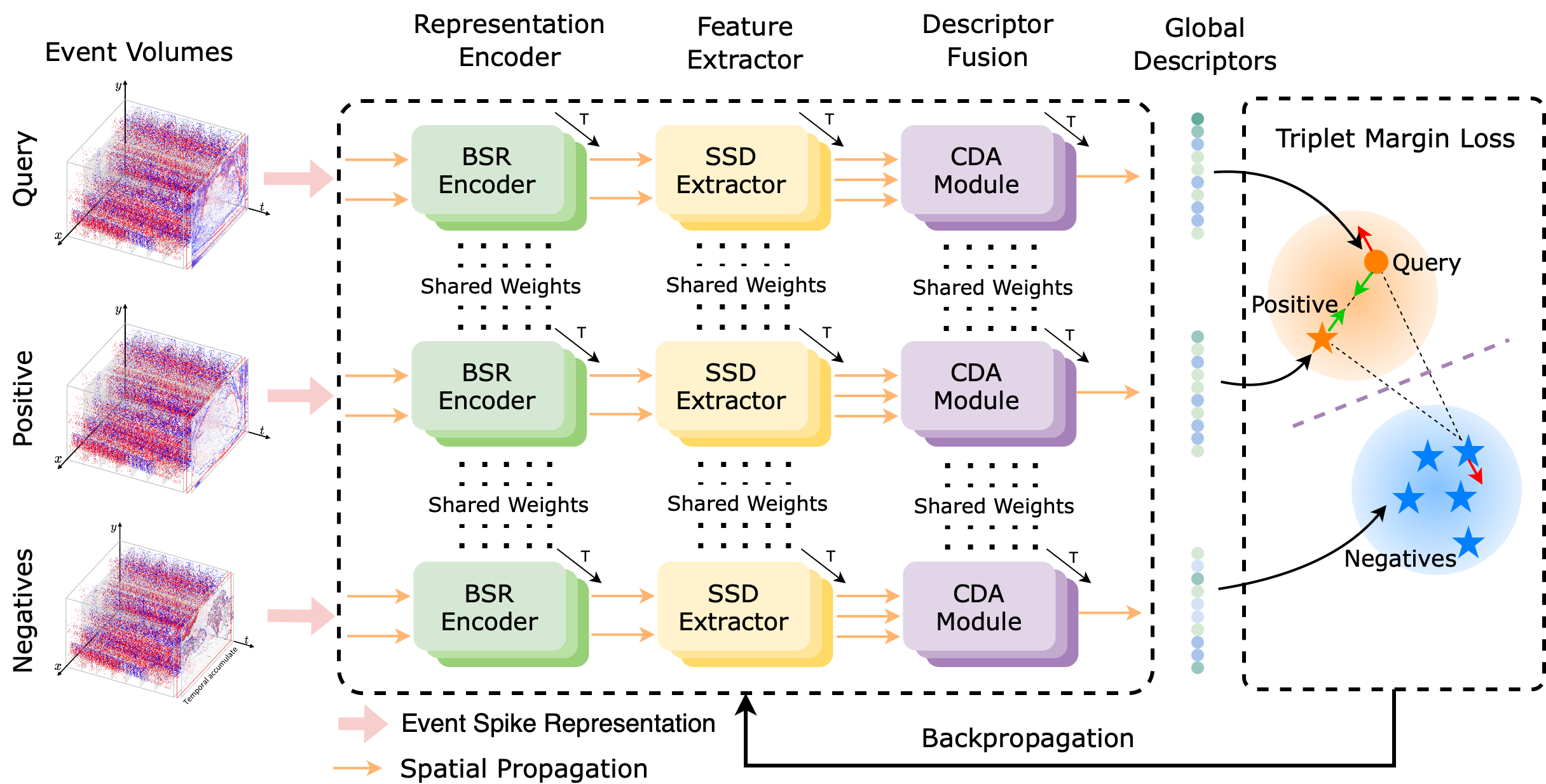}
\caption{Overview of our Spike-EVPR pipeline.}
\label{fig:intro}
\end{figure}
Despite the natural compatibility between SNNs and event streams, two key challenges remain in applying SNNs to EVPR.
First, regarding data representation, directly converting raw event streams into temporally dense spike tensors introduces excessive and redundant time steps~\cite{fischer2020event,kong2022event}. This redundancy leads to inefficient spike activity and impedes the learning of meaningful representations. 
Therefore, it is critical to design spike-compatible representations that preserve essential spatio-temporal cues while reducing temporal redundancy, enabling more effective SNN training.
Second, regarding network architecture, while deeper SNNs have been explored for generic vision tasks~\cite{fang2021deep, yang2023sa, gu2020tactilesgnet}, their application in EVPR remains underexplored. 
Specifically, there is a lack of effective SNN architectures capable of processing spatio-temporal cues to extract discriminative global descriptors, highlighting the critical need for specialized deep SNN pipelines tailored specifically to EVPR.

To address these challenges, we introduce Spike-EVPR, a spiking neural network framework specifically designed for EVPR, as illustrated in Fig.~\ref{fig:intro}. 
Spike-EVPR is built on two key ideas.
First, instead of directly feeding ANN-based event representations with redundant time steps into SNNs,
we design two complementary spike-compatible event representations, the Multi-Channel Spike Tensor (MCS-Tensor) and the Time-Surface Spike Tensor (TSS-Tensor).
These representations reduce temporal redundancy and alleviate spatial sparsity in event streams, providing suitable spike inputs for SNN training.
Second, we construct a deep spiking residual architecture for EVPR. The Bifurcated Spike Residual Encoder (BSR-Encoder) extracts complementary spatio-temporal spike features from the two representations. 
These features are further decomposed by the Shared and Specific Descriptor Extractor (SSD-Extractor) into shared and representation-specific components, yielding sub-descriptors with distinct characteristics. 
Finally, the Cross-Descriptor Aggregation Module (CDA-Module) learns the importance of each sub-descriptor and integrates them into a discriminative global descriptor for place recognition. 
Extensive experiments demonstrate that Spike-EVPR achieves state-of-the-art performance while significantly reducing energy consumption. Our contributions are summarized as follows:
\begin{itemize}
    \item We propose Spike-EVPR, a directly trained, end-to-end SNN framework specifically designed for EVPR, enabling energy-efficient and robust place recognition from sparse and asynchronous event streams.
    \item We introduce two complementary spike-compatible event representations, MCS-Tensor and TSS-Tensor, which jointly provide efficient and informative spatio-temporal spike inputs for SNN learning in EVPR.
    \item We design a deep spiking residual architecture consisting of the BSR-Encoder, SSD-Extractor, and CDA-Module, enabling feature extraction, decomposition, and aggregation for discriminative global descriptors.
    \item We achieve state-of-the-art results on Brisbane-Event-VPR and DDD20, with Recall@1 gains of 7.61\% and 13.20\%, while significantly reducing energy consumption compared with ANN-based EVPR baselines.
\end{itemize}

\section{Related Works}
\subsection{Event-based Visual Place Recognition}
Motivated by the unique spatio-temporal characteristics of event cameras, numerous methods have been proposed for event-based visual place recognition (EVPR).
Ensemble-Event-VPR~\cite{fischer2020event} presents an ensemble EVPR framework that aggregates predictions from multiple reconstruction methods, feature extractors, and temporal resolutions.
EventVLAD~\cite{lee2021eventvlad} introduces edge images generated from event streams into the NetVLAD pipeline.
Event-VPR~\cite{kong2022event} converts events into EST voxel grids and extracts descriptors with a deep residual network followed by a VLAD layer. 
Sparse-Event-VPR~\cite{fischer2022many} improves efficiency by using only sparse key events for descriptor extraction. 
VEFNet~\cite{huang2022vefnet} and FE-Fusion-VPR~\cite{hou2023fe} combine frame and event information through attention-based fusion, while EFormer-VPR~\cite{zhang2024eformer} employs transformer-based networks to extract features from both modalities and fuses them through a scoring module.
Recently, Flash~\cite{ramanathan2025prepare} departs from dense event representations and operates directly on active pixel locations, enabling ultra-low-latency place recognition under extreme motion.
Despite these advances, existing EVPR pipelines are constrained by conventional ANN architectures, which typically convert events into dense frame-like representations and thus fail to exploit the inherent sparsity and asynchronous dynamics of raw event streams. 
This motivates the development of spiking-based EVPR frameworks that more naturally align with the event generation process.

\subsection{Spiking Neural Network}
The sparse and asynchronous nature of event cameras aligns naturally with spiking neural networks (SNNs), making them a promising paradigm for event-driven vision tasks. SNNs have shown effectiveness across optical flow prediction~\cite{yang2023sa}, motion segmentation~\cite{parameshwara2021spikems}, object detection~\cite{gu2020tactilesgnet}, and depth estimation~\cite{wu2022mss}. 
Several recent works explore SNNs for EVPR.
Ev-ReconNet~\cite{lee2023ev} employs an ANN-to-SNN network to reconstruct edge images from events before applying NetVLAD for descriptor extraction. NeuroGPR~\cite{yu2023brain} uses an SNN model to process event data for robot place recognition.
In parallel, LENS~\cite{hines2025compact} presents a fully neuromorphic VPR system that enables real-time, energy-efficient localization on robotic platforms.
Despite these advances, existing SNN-based EVPR methods still rely on conventional ANN components for key stages such as image reconstruction, or are limited by extremely compact, hardware-specific models, which limits their scalability to large-scale VPR benchmarks. This motivates exploring more general and scalable spiking architectures for EVPR.

\section{Methodology}
\subsection{Overview}
\begin{figure*}[t]
\centering
\includegraphics[width=\textwidth]{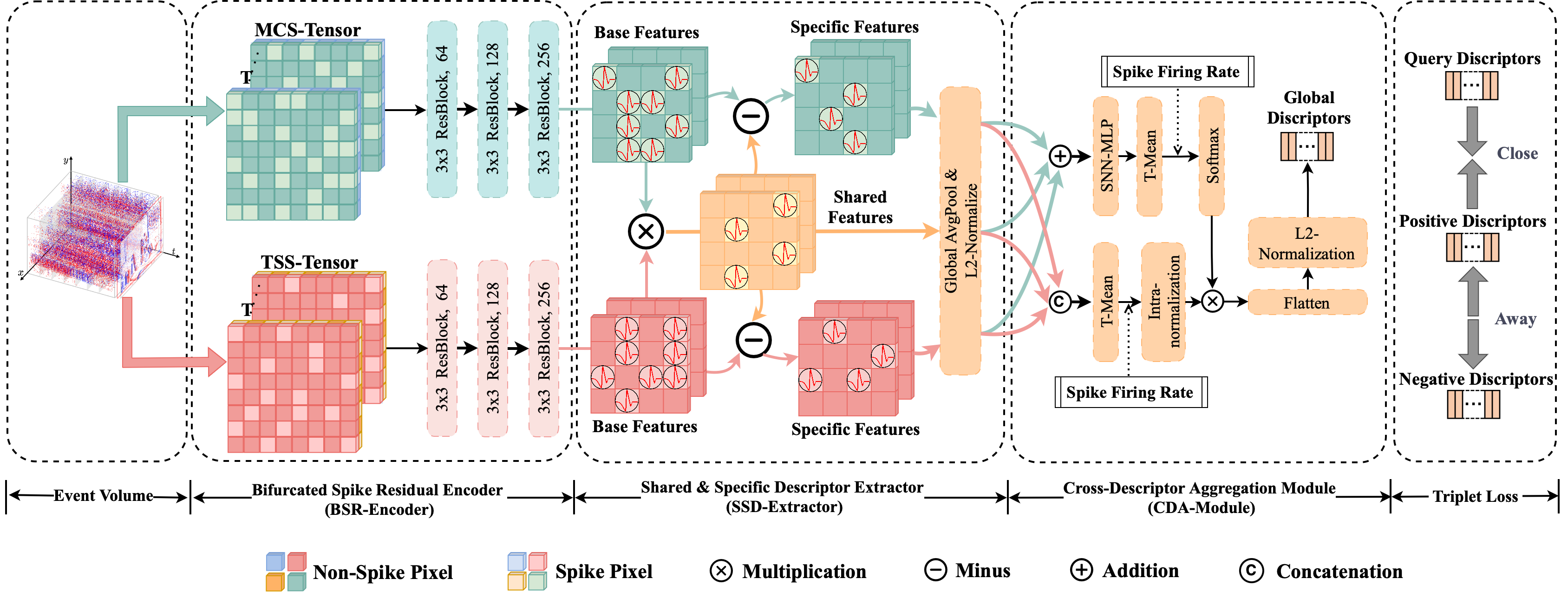}
\caption{Architecture of Spike-EVPR. The event volume is first converted into twon spike-compatible representations: MCS-Tensor and TSS-Tensor. These are encoded by the Bifurcated Spike Residual Encoder (BSR-Encoder) into feature maps. 
The Shared \& Specific Descriptor Extractor (SSD-Extractor) then generates three complementary sub-descriptors, which are subsequently aggregated by Cross-Descriptor Aggregation Module (CDA-Module) to produce the robust global descriptor.}
\vspace{-3pt}
\label{fig:arch}
\end{figure*}

The architecture of the proposed Spike-EVPR is shown in Fig.~\ref{fig:arch}. We first converts event volumes into two spike-compatible representations: the Multi-Channel Spike Tensor (MCS-Tensor) and the Time-Surface Spike Tensor (TSS-Tensor), which compactly encode spatio-temporal information for efficient training. 
Based on these, we introduce three key sub-modules for robust global descriptor extraction. 
First, the Bifurcated Spike Residual Encoder (BSR-Encoder), based on deep spiking residual blocks, extracts spike features from both representations.
Next, the Shared \& Specific Descriptor Extractor (SSD-Extractor) splits and recombines these features to produce sub-descriptors with complementary characteristics. 
Finally, the Cross-Descriptor Aggregation Module (CDA-Module) fuses the multiple sub-descriptors generated by the SSD-Extractor, adaptively learning their weights and aggregating them into a refined global descriptor.
The entire pipeline is directly trained end-to-end via weakly supervised triplet contrastive learning.

\subsection{SNN-Tailored Event Representations}
\begin{figure*}[t]
\centering
\includegraphics[width=0.9\textwidth]{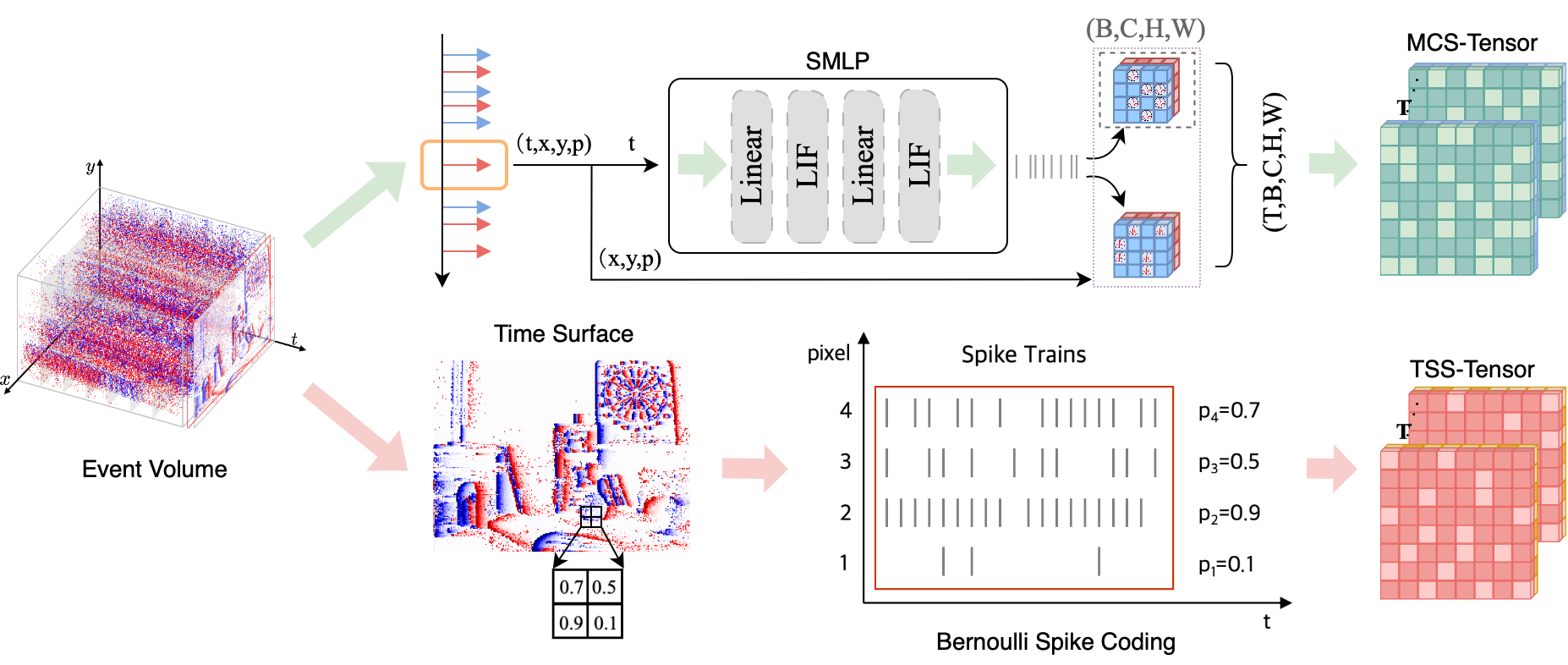}
\caption{Detailed illustration of our SNN-Tailored event representations.}
\label{fig:representation}
\vspace{-3pt}
\end{figure*}

Event volumes in EVPR suffer from two fundamental issues:
extremely high temporal resolution, which leads to redundant spike time steps, and severe spatial sparsity, which makes it difficult for SNNs to recover local structure. A single spike representation is therefore insufficient, since reducing temporal redundancy tends to lose spatial detail, whereas densifying spatial information often ignores the fine-grained temporal cues that make events valuable.
To address these complementary challenges, we construct two spike-compatible representations that focus on different aspects of the event stream and use them jointly in our network:
the Multi-Channel Spike Tensor (MCS-Tensor) and the Time-Surface Spike Tensor (TSS-Tensor), as shown in Fig~\ref{fig:representation}.

\subsubsection{Multi-Channel Spike Tensor (MCS-Tensor)}
The Multi-Channel Spike Tensor (MCS-Tensor) is a learnable spike representation designed to address excessive  and redundant time steps in temporally dense event streams by adaptively mapping events to appropriate time steps within the spiking network.
At its core, MCS-Tensor encodes event timestamps using a Spiking Multi-Layer Perceptron (SMLP), enabling temporally compact and information-preserving spike representations, denoted as $\boldsymbol{f}_\text{SMLP}(\cdot)$:
\begin{equation}
\label{eq:smlp}
\begin{aligned}
\boldsymbol{f}_\text{SMLP}(\cdot)
&=\boldsymbol{f}_\text{LIF,2}\left(\boldsymbol{f}_{\text{FC,2}}\left(\boldsymbol{f}_\text{LIF,1}\left(\boldsymbol{f}_{\text{FC,1}}\left(\boldsymbol{E}\right)\right)\right)\right),
\end{aligned}
\end{equation}
where $\boldsymbol{f}_{\text{FC},1}(\cdot)$ and $\boldsymbol{f}_{\text{FC},2}(\cdot)$ are the two fully connected layers, respectively. Each fully connected layer is associated with the LIF neuron layer $\boldsymbol{f}_{\text{LIF}}(\cdot)$ for the spike coding of event timestamps.
This encoding transforms the event timestamps into appropriate spike time steps and yields a spiking tensor with dimensions $\mathbb{R}^{\mathrm{T}\times\mathrm{C}\times\mathrm{H}\times\mathrm{W}}$, which serves as input to the subsequent feature extraction pipeline. 
The MCS-Tensor is defined as follows:
\begin{equation}
% \label{eq:mcstensor}
\begin{aligned}
\boldsymbol{\mathcal{V}}_{\pm}(x^{\prime}, y^{\prime}, t^{\prime})
&=\sum_{e_i \in \boldsymbol{E}_\pm } \left|p_i\right|\boldsymbol{\delta}\left(x^{\prime}-x_i,y^{\prime}-y_i\right) \boldsymbol{f}_\text{SMLP}\left(t^{\prime}-t_i^*\right),
\end{aligned}
\end{equation}
where $x^{\prime}, y^{\prime}, t^{\prime}$ denote the space-time coordinates of the spiking tensor, with $x^{\prime} \in [0, \mathrm{W}-1]$, $y^{\prime} \in [0, \mathrm{H}-1]$, and $t^{\prime} \in [0, \mathrm{T}-1]$, where $(\mathrm{W}, \mathrm{H})$ is the spatial resolution, and $\mathrm{T}$ is the number of discrete time steps. The normalized timestamp $t_i^* =\left(\mathrm{T}-1\right) \frac{\left(t_i-t_1\right)}{\left(t_\text{last}-t_1\right)}$ means event timestamps are normalized and distributed to the corresponding time steps.

\subsubsection{Time-Surface Spike Tensor (TSS-Tensor)}
To complement MCS-Tensor and further address the spatial sparsity of event data, we propose the Time-Surface Spike Tensor (TSS-Tensor), a spike-compatible representation designed to preserve local spatial structure while encoding temporal dynamics around event locations. 
Inspired by the concept of time surface (TS) ~\cite{lagorce2016hots}, we construct a dense map where each pixel records the timestamp of the most recent event using an exponential decay kernel, capturing fine-grained temporal context. To convert this into a discrete spike tensor suitable for SNNs, we apply Bernoulli sampling to the decayed values. 
The TS map is defined as follows:
\begin{equation}
\label{eq:Ts}
\begin{aligned}
\boldsymbol{\mathcal{\tilde{T}}}_{\pm}(x^{\prime},y^{\prime})=\exp{\left(-\frac{t_\text{end}-t_{\text{last},\pm}(x^{\prime},y^{\prime})}{\eta}\right)},
\end{aligned}
\end{equation}
where $t_\text{end}$ is the end time of the event volume, $t_{\text{last}, \pm}(x^{\prime},y^{\prime})$ denotes the timestamp of the most recent positive and negative events at pixel $(x^{\prime},y^{\prime})$, and $\eta$ is the decay rate. 
We assume that the firing rate of the encoded spike trains is correlated with the TS-map at each pixel. 
To convert the continuous TS map into a discrete spike tensor, we apply Bernoulli sampling over $\mathrm{T}$ time steps, as follows:
\begin{equation}
\label{eq:tsstensor}
\begin{aligned}
\boldsymbol{\mathcal{T}}_{\pm}(x^{\prime},y^{\prime},t^{\prime})\sim\mathcal{B}(\boldsymbol{\mathcal{\tilde{T}}}_{\pm},p_{x^{\prime},y^{\prime}},\mathrm{T}),
\end{aligned}
\end{equation}
where $\mathcal{B}(\cdot)$ denotes the Bernoulli distribution, $p_{x^{\prime},y^{\prime}}$ is the probability at pixel $(x^{\prime},y^{\prime})$ derived from the TS map $\boldsymbol{\mathcal{\tilde{T}}}(x^{\prime}, y^{\prime})$, and $\mathrm{T}$ is the number of sampling steps. 
By recovering local spatial context and encoding temporal cues in spike form, this process complements MCS-Tensor and yields a compact spatio-temporal representation for SNNs.

\subsection{Deep Spiking Network Architecture}
\subsubsection{Bifurcated Spike Residual Encoder (BSR-Encoder)}
Based on the dual SNN-tailored event representations described above, we design a spiking residual network with powerful feature extraction capabilities for robust place recognition. 
As shown in Fig.~\ref{fig:arch}, our proposed Bifurcated Spike Residual Encoder (BSR-Encoder) extracts the initial feature maps from the two spike-compatible representations using a dual-stream deep spiking residual network, which can be summarized as follows:
\begin{equation}
\begin{aligned}
\boldsymbol{M}_\text{MCS}, \boldsymbol{M}_\text{TSS}
&= \boldsymbol{f}_{\text{BSR-Encoder}}(\boldsymbol{\mathcal{V}}_{\pm}, \boldsymbol{\mathcal{T}}_{\pm}) \\
&= \boldsymbol{f}_\text{SR13-MCS}(\boldsymbol{\mathcal{V}}_{\pm}), \boldsymbol{f}_\text{SR13-TSS}(\boldsymbol{\mathcal{T}}_{\pm}),
\end{aligned}
\end{equation}
where $\boldsymbol{f}_{\text{SR13-MCS}}(\cdot)$ and $\boldsymbol{f}_{\text{SR13-TSS}}(\cdot)$ denote 13-layer spiking residual networks adapted from SEW-ResNet18~\cite{fang2021deep}, used to process $\boldsymbol{\mathcal{V}}_{\pm}$ (MCS-Tensor) and $\boldsymbol{\mathcal{T}}_{\pm}$ (TSS-Tensor), respectively. 
The resulting feature maps $\boldsymbol{M}_{\text{MCS}}, \boldsymbol{M}_{\text{TSS}}\in\mathbb{B}^{\mathrm{T}\times\mathrm{C}\times\mathrm{H}\times\mathrm{W}}$ retain fine-grained spatio-temporal patterns essential for downstream place recognition tasks. 
We construct the BSR-Encoder using spiking residual blocks, as follows:
\begin{equation}
\begin{aligned}
\boldsymbol{F}_{n}(\cdot) &= \boldsymbol{f}_{\text{BN},n}(\boldsymbol{f}_{\text{Conv},n}(\boldsymbol{f}_{\text{LIF1},n}(\boldsymbol{f}_{\text{BN},n}(\boldsymbol{f}_{\text{Conv},n}(\cdot))))), \\
\boldsymbol{O}_{n}^{t} &= \boldsymbol{f}_{\text{LIF2},n}(\boldsymbol{F}_{n}(\boldsymbol{O}_{n-1}^{t}) + \boldsymbol{O}_{n-1}^{t}),
\end{aligned}
\end{equation}
where $\boldsymbol{F}_{n}(\cdot)$ denotes the residual block, consisting of a $3\times3$ convolution $\boldsymbol{f}_{\text{Conv},n}(\cdot)$, a batch normalization layer $\boldsymbol{f}_{\text{BN},n}(\cdot)$, and a LIF neuron $\boldsymbol{f}_{\text{LIF},n}(\cdot)$. 
The block output is added to the input $\boldsymbol{O}_{n-1}^{t}$ to produce the output spike trains at timestep $t$. 
This structure preserves temporal dynamics and stabilizes deep SNN training by mitigating gradient vanishing and explosion, as shown in~\cite{fang2021deep}, where residual connections enhance gradient flow in deep spiking architectures.

\subsubsection{Shared \& Specific Descriptor Extractor (SSD-Extractor)}
To fully exploit the complementary characteristics of the initial feature maps extracted from the two spike representations, we design the Shared \& Specific Descriptor Extractor (SSD-Extractor). This module enhances descriptor discriminability by explicitly modeling both shared and specific features in the outputs of the BSR-Encoder. 
As illustrated in Fig.~\ref{fig:arch}, the initial spike feature maps are binary (0/1). We compute the shared feature map via element-wise multiplication of the two feature maps, highlighting the common activated regions. The specific feature maps for each representation are then obtained by subtracting the shared feature map from the corresponding initial feature map, thereby isolating the modality-specific features. 
These operations can be formulated as follows:
\begin{equation}
\begin{aligned}
\boldsymbol{X}_1 &= \boldsymbol{M}_{\text{MCS}} \otimes \boldsymbol{M}_{\text{TSS}}, \\
\boldsymbol{X}_2 &= \boldsymbol{M}_{\text{MCS}} \ominus \boldsymbol{X}_1, \\
\boldsymbol{X}_3 &= \boldsymbol{M}_{\text{TSS}} \ominus \boldsymbol{X}_1,
\end{aligned}
\end{equation}
where $\boldsymbol{X}_1\in \mathbb{B}^{\mathrm{T}\times \mathrm{C}\times \mathrm{H}\times \mathrm{W}}$ denotes the shared feature map between both representations. 
$\boldsymbol{X}_2$ and $\boldsymbol{X}_3 \in\mathbb{B}^{\mathrm{T}\times\mathrm{C}\times\mathrm{H}\times\mathrm{W}}$ correspond to the specific feature maps of the MCS-Tensor and TSS-Tensor, respectively. 
Here, $\otimes$ and $\ominus$ denote element-wise multiplication and subtraction operations. 

Following that, we apply global average pooling (GAP) followed by L2 normalization to the shared feature map $\boldsymbol{X}_1$ and the specific feature maps $\boldsymbol{X}_2, \boldsymbol{X}_3$, yielding three sub-descriptors $\boldsymbol{D}_1,\boldsymbol{D}_2,\boldsymbol{D}_3\in\mathbb{R}^{\mathrm{T}\times\mathrm{C}}$, defined as:
\begin{equation}
\begin{aligned}
\boldsymbol{D}_1, \boldsymbol{D}_2, \boldsymbol{D}_3
&= \boldsymbol{f}_{\text{L2-norm}}(\boldsymbol{f}_\text{GAP}(\boldsymbol{X}_1, \boldsymbol{X}_2, \boldsymbol{X}_3)),
\end{aligned}
\end{equation}
Among them, $\boldsymbol{D}_1$ captures shared information, while $\boldsymbol{D}_2$ and $\boldsymbol{D}_3$ encode information specific to the MCS-Tensor and TSS-Tensor branches, respectively.

\subsubsection{Cross-Descriptor Aggregation Module (CDA-Module)}
In the SSD-Extractor, we obtain shared and specific sub-descriptors from the two spike representations. 
To effectively fuse them, we propose the Cross-Descriptor Aggregation Module (CDA-Module) $\boldsymbol{f}_\text{CDA-Module}(\cdot)$, which learns to selectively emphasize salient information from different sub-descriptors and produce a robust global descriptor:
\begin{equation}
\begin{aligned}
\tilde{\boldsymbol{D}} = \boldsymbol{f}_\text{CDA-Module}(\boldsymbol{D}_1, \boldsymbol{D}_2, \boldsymbol{D}_3).
\end{aligned}
\end{equation}
We first learn the weights of the three sub-descriptors (one shared and two specific). 
The descriptors $\boldsymbol{D}_1, \boldsymbol{D}_2, \boldsymbol{D}_3$ are element-wise summed to form an aggregated descriptor $\boldsymbol{D_\text{add}}\in \mathbb{R}^{\mathrm{T}\times \mathrm{C}}$. 
This descriptor is then passed through a Spiking Multi-Layer Perceptron (SMLP) to produce $\boldsymbol{D}^{\prime}_\text{add} \in \mathbb{R}^{\mathrm{T}\times \mathrm{3}}$, where 3 denotes the number of sub-descriptors. 
$\boldsymbol{D}^{\prime}_\text{add}$ is then averaged along the temporal dimension to compute the spike firing rates, which are normalized via a softmax function to generate the final descriptor weights, $\boldsymbol{w} \in \mathbb{R}^{\mathrm{3}}$:
\begin{equation}
\begin{aligned}
\boldsymbol{D_\text{add}} &= \boldsymbol{D_1} \oplus \boldsymbol{D_2} \oplus \boldsymbol{D_3}, \\
\boldsymbol{D}^{\prime}_\text{add} &= \boldsymbol{f}_{\text{FC},2}\left(\boldsymbol{f}_\text{LIF}\left(\boldsymbol{f}_{\text{FC},1}\left(\boldsymbol{D}_\text{add}\right)\right)\right), \\
\boldsymbol{w} &= \boldsymbol{f}_\text{SM}\left(\boldsymbol{f}_\text{Mean}\left(\boldsymbol{D}^{\prime}_\text{add}\right)\right)
\end{aligned}
\end{equation}
where $\boldsymbol{f}_{\text{FC},1}(\cdot):\mathbb{R}^{\mathrm{T}\times256\times1}\rightarrow\mathbb{R}^{\mathrm{T}\times\mathrm{M}\times1}$ and $\boldsymbol{f}_{\text{FC},2}(\cdot):\mathbb{R}^{\mathrm{T}\times\mathrm{M}\times1}\rightarrow\mathbb{R}^{\mathrm{T}\times3\times1}$ are two fully connected layers, with $\mathrm{M}=64$ denoting the hidden dimension of the SMLP.
$\boldsymbol{f}_\text{Mean}(\cdot):\mathbb{R}^{\mathrm{T}\times 3\times1}\rightarrow\mathbb{R}^{3\times1}$ computes average spike firing rates along the temporal dimension, and $\boldsymbol{f}_\text{SM}(\cdot)$ applies softmax to normalize the learned descriptors weights. 

Secondly, the three sub-descriptors $\boldsymbol{D}_{1},\boldsymbol{D}_{2},\boldsymbol{D}_{3}$ are concatenated along a new axis to form a unified tensor $\boldsymbol{D}_{\text{concat}} \in \mathbb{R}^{\mathrm{T} \times3\times\mathrm{C}}$. 
We then compute the average spike firing rates over the temporal dimension, yielding $\boldsymbol{D}^{\prime}_{\text{concat}} \in \mathbb{R}^{3 \times\mathrm{C}}$, followed by intra-normalization along the channel dimension:
\begin{equation}
\begin{aligned}
&\boldsymbol{D}_\text{concat}= \mathop{\Big{\|}}_{i=1}^3(\boldsymbol{D}_{i}),\\
&\boldsymbol{D}^{\prime}_\text{concat} =\boldsymbol{f}_\text{Intra-norm}\left(\boldsymbol{f}_{\text{Mean}}(\boldsymbol{D}_\text{concat})\right),
\end{aligned}
\end{equation}
where $\mathop{\Big{\|}} _{i=1}^{3}(\cdot)$ denotes the concatenation operation, and $\boldsymbol{f}_\text{Intra-norm}(\cdot)$ refers to the intra-normalization operation. 

Finally, the weight vector $\boldsymbol{w}$ is applied to the concatenated descriptor $\boldsymbol{D}^{\prime}_{\text{concat}}$ via element-wise multiplication. The resulting tensor is flattened and L2-normalized to obtain the final refined global descriptor $\boldsymbol{D}^{\prime} \in \mathbb{R}^{3\mathrm{C}}$, defined as: 
\begin{equation}
\begin{aligned}
\tilde{\boldsymbol{D}} = \boldsymbol{f}_\text{L2-norm}\left(\boldsymbol{f}_{\text{Flatten}}(\boldsymbol{D}_\text{concat}^{\prime} \otimes \boldsymbol{w})\right),
\end{aligned}
\end{equation}
where $\boldsymbol{f}_\text{Flatten}(\cdot): \mathbb{R}^{3\times\mathrm{C}} \rightarrow \mathbb{R}^{3\mathrm{C}}$ denotes the flattening operation.
This weighted aggregation enables the global descriptor to adaptively integrate complementary cues from the shared and specific sub-descriptors, thereby enhancing both its discriminative power and generalization capability.

\subsection{Training Loss}
\iffalse
NOTE TO Co-authors:
Please do not move the following table.
It's intentionally placed here for layout reason.
\fi

To train our proposed framework, we adopt the triplet loss $\mathcal{L}_\text{triplet}$ introduced in~\cite{arandjelovic2016netvlad} for weakly supervised training. 
The objective is to learn a representation space where the distance between the query $\boldsymbol{D}_\mathrm{q}$ and the positive sample $\boldsymbol{D}_\mathrm{pos}$ is minimized, while the distance to hard negatives $\boldsymbol{D}_\mathrm{neg}$ is maximized. The loss is defined as:
\begin{equation}
\begin{aligned}
\mathcal{L}_\text{triplet}= \sum_{i=1}^N \left[\|\boldsymbol{D}_\mathrm{q} - \boldsymbol{D}_\mathrm{pos}\|_2^2 
- \|\boldsymbol{D}_\mathrm{q} - \boldsymbol{D}_\mathrm{neg}^{(i)}\|_2^2 +
m \right]_+,
\end{aligned}
\end{equation}
where $[\cdot]_+ = \max(\cdot,0)$ denotes the hinge loss, ${N}$ is the number of triplets in a batch, and ${m}$ is the margin hyperparameter. 
This loss promotes globally discriminative descriptors by pulling positive pairs closer and pushing hard negatives apart in the learned feature space.

\section{Experiments}
\subsection{Experimental Setup}
\subsubsection{Datasets}
To evaluate the performance of our method, we conduct experiments on two EVPR datasets: Brisbane-Event-VPR~\cite{fischer2020event} and DDD20~\cite{hu2020ddd20}. The Brisbane-Event-VPR dataset contains recordings from a DAVIS camera and GPS, covering six 8 km paths at different times of the day. We focus on four sequences: sunrise, morning, daytime, and sunset. The DDD20 dataset is a comprehensive driving dataset captured by event cameras under various lighting conditions. We select six sequences from two urban scenes, including two with glare and three from highway scenarios.

\subsubsection{Implementation Details}
We train our model using triplet ranking loss as weak supervision. During evaluation, a query is considered accurate if the retrieved image is within a 75m radius of the true GPS location. The event volume time intervals are set to 0.25s for Brisbane-Event-VPR and 0.2s for DDD20. For consistency, we use the same settings across experiments, with a learning rate of 0.001, a training batch size of 2, and a cache batch size of 100. All experiments are conducted using an NVIDIA RTX 4090.

\subsubsection{Evaluation Metrics}
We adopt Precision-Recall (PR) curves and Recall@N as evaluation metrics, which are widely used in place recognition tasks~\cite{hou2023fe, kong2022event}. 
Additionally, we report F1-max following~\cite{hou2023fe}, which captures the best trade-off between precision and recall across thresholds.

\subsection{Comparisons with the State-of-the-Art Methods}
\begin{table*}[t]
\centering
\caption{Quantitative results on Brisbane-Event-VPR and DDD20 datasets. \textbf{Bold} is best.
\textit{Note:} ``---'' indicates missing results due to (i) no reported DDD20 performance or (ii) insufficient code or implementation details for reproduction.}
\setlength{\tabcolsep}{0.006\linewidth}
{
\renewcommand{\arraystretch}{1.2}
\setlength{\tabcolsep}{0.006\linewidth}
\begin{tabular}{c|c|cccc|c|cc|cp{1cm}}
\toprule[.05cm]
\multirow{4}{*}{Models}
&\multirow{4}{*}{\makecell{EVPR Methods}}
&\multicolumn{5}{c|}{\makecell{Brisbane-Event-VPR~\cite{fischer2020event} Dataset}}
&\multicolumn{3}{c}{\makecell{DDD20~\cite{hu2020ddd20} Dataset}}  \\
\cline{3-10}
&
&\multicolumn{4}{c|}{\multirow{2}{*}{\makecell{Recall@1/5 (\%) $\uparrow$ \\ F1-max $\uparrow$}}}
&\multirow{3}{*}{\makecell{Avg\\Recall@1\\(\%)$\uparrow$}}
&\multicolumn{2}{c|}{\multirow{2}{*}{\makecell{Recall@1/5 (\%) $\uparrow$ \\ F1-max $\uparrow$}}}
&\multirow{3}{*}{\makecell{Avg\\Recall@1\\(\%)$\uparrow$}} \\
&   &\multicolumn{4}{c|}{}   &   &\multicolumn{2}{c|}{}   &   \\
\cline{3-6}\cline{8-9}
&   &ss1 / ss2  &ss1 / sr   &ss1 / mn   &ss1 / dt   &   &**83 / **48    &**81 / **68    &   \\
\midrule
\multirow{6}{*}{ANN}
&\multirow{2}{*}{\makecell{Ensemble-Event-VPR~\cite{fischer2020event}}}
&\multirow{2}{*}{\makecell{87.33, 95.70 \\ 0.9345}}
&\multirow{2}{*}{\makecell{58.82, 82.42 \\ 0.7246}}
&\multirow{2}{*}{\makecell{58.42, 78.33 \\ 0.7550}}
&\multirow{2}{*}{\makecell{43.62, 62.71 \\ 0.6319}}
&\multirow{2}{*}{\makecell{62.05}}
&\multirow{2}{*}{\makecell{32.02, 50.17 \\ 0.4967}}
&\multirow{2}{*}{\makecell{ 7.84, 14.99 \\ 0.1465}}
&\multirow{2}{*}{\makecell{19.93}}   \\
&   &   &   &   &   &   &   &   &   \\
\cline{2-10}
&\multirow{2}{*}{\makecell{Event-VPR~\cite{kong2022event}}}
&\multirow{2}{*}{\makecell{84.79, 93.83 \\ 0.9236}}
&\multirow{2}{*}{\makecell{65.65, 86.52 \\ 0.7984}}
&\multirow{2}{*}{\makecell{66.67, 84.26 \\ 0.7946}}
&\multirow{2}{*}{\makecell{44.54, 66.10 \\ 0.6705}}
&\multirow{2}{*}{\makecell{65.41}}
&\multirow{2}{*}{\makecell{43.89, 66.67 \\ 0.6057}}
&\multirow{2}{*}{\makecell{8.52, 21.33 \\ 0.1578}}
&\multirow{2}{*}{\makecell{26.21}}   \\
&   &   &   &   &   &   &   &   &   \\
\cline{2-10}
&\multirow{2}{*}{\makecell{Sparse-Event-VPR~\cite{fischer2022many}}}
&\multirow{2}{*}{\makecell{\textbf{96.35}, \textbf{98.92} \\ \textbf{0.9832}}}
&\multirow{2}{*}{\makecell{79.41, 92.51 \\ 0.8844}}
&\multirow{2}{*}{\makecell{69.76, 87.79 \\ 0.8162}}
&\multirow{2}{*}{\makecell{47.13, 69.52 \\ 0.6427}}
&\multirow{2}{*}{\makecell{73.16}}
&\multirow{2}{*}{\makecell{---}}
&\multirow{2}{*}{\makecell{---}}
&\multirow{2}{*}{\makecell{---}}   \\
&   &   &   &   &   &   &   &   &   \\
\cline{1-10}
\multirow{4}{*}{SNN}
&\multirow{2}{*}{\makecell{Ev-ReconNet~\cite{lee2023ev}}}
&\multirow{2}{*}{\makecell{95.73, 98.26 \\ 0.9573}}
&\multirow{2}{*}{\makecell{66.08, 86.72 \\ 0.6608}}
&\multirow{2}{*}{\makecell{44.60, 66.70 \\ 0.4460}}
&\multirow{2}{*}{\makecell{36.37, 58.33 \\ 0.3637}}
&\multirow{2}{*}{\makecell{60.70}}
&\multirow{2}{*}{\makecell{---}}
&\multirow{2}{*}{\makecell{---}}
&\multirow{2}{*}{\makecell{---}}   \\
&   &   &   &   &   &   &   &   &   \\
\cline{2-10}\rule{0pt}{8pt}
&\multirow{2}{*}{\makecell{\textbf{Spike-EVPR (Ours)}}}
&\multirow{2}{*}{\makecell{87.78, 96.04 \\ 0.9227}}
&\multirow{2}{*}{\makecell{\textbf{86.56}, \textbf{95.27} \\ \textbf{0.9289}}}
&\multirow{2}{*}{\makecell{\textbf{88.98}, \textbf{96.33} \\ \textbf{0.9423}}}
&\multirow{2}{*}{\makecell{\textbf{59.77}, \textbf{79.45} \\ \textbf{0.6427}}}
&\multirow{2}{*}{\makecell{\textbf{80.77}}}
&\multirow{2}{*}{\makecell{\textbf{58.12}, \textbf{79.76} \\ \textbf{0.7118}}}
&\multirow{2}{*}{\makecell{\textbf{20.69}, \textbf{37.07} \\ \textbf{0.4419}}}
&\multirow{2}{*}{\makecell{\textbf{39.41}}}   \\
&   &   &   &   &   &   &   &   &   \\
\bottomrule[.05cm]
\end{tabular}
}
\label{tab:compare}
\vspace{-3pt}
\end{table*}

\begin{figure*}[htbp]
\centering
\includegraphics[width=0.95\textwidth]{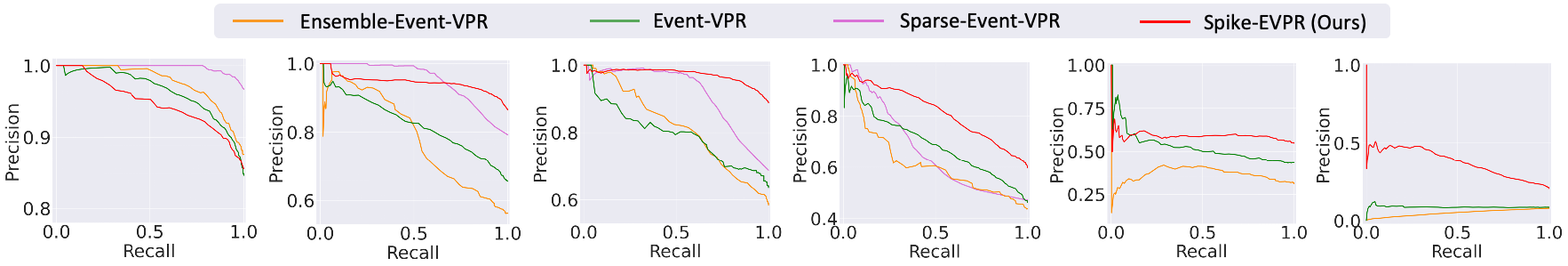}
\includegraphics[width=0.95\textwidth]{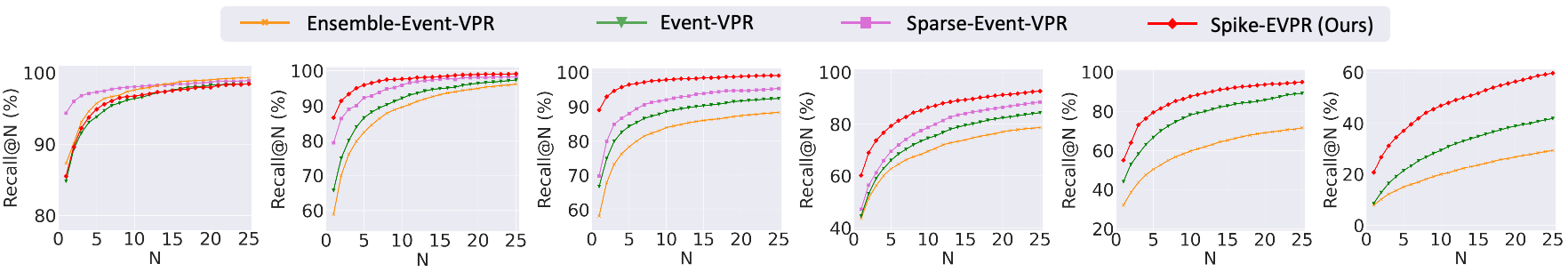}
\caption{PR curves and Recall@N comparisons on Brisbane-Event-VPR (first four) and DDD20 (last two) datasets. Our Spike-EVPR (in red) consistently outperforms existing SOTA EVPR methods across most scenes.}
\label{fig:qualitative}
\vspace{-8pt}
\end{figure*}
We compare Spike-EVPR with state-of-the-art (SOTA) ANN-based approaches, including Ensemble-Event-VPR~\cite{fischer2020event}, Event-VPR~\cite{kong2022event}, and Sparse-Event-VPR~\cite{fischer2022many}, as well as the SNN-based method Ev-ReconNet~\cite{lee2023ev}. 
As shown in Tab.~\ref{tab:compare}, our method achieves superior performance on both the Brisbane-Event-VPR and DDD20 datasets, with an average Recall@1 gain of 7.61\% over Sparse-Event-VPR, and up to 19.48\% and 13.20\%  over Ensemble-Event-VPR and Event-VPR on DDD20. 
The PR curves in Fig.~\ref{fig:qualitative} further verify our advantage, highlighting the benefits of efficient spatio-temporal encoding and cross-representation aggregation via the SSD-Extractor and CDA-Module. These modules extract shared, specific, and weighted features to generate scene descriptors, addressing the challenge of extracting discriminative representations from asynchronous event streams. 

Ev-ReconNet, a representative SNN-based EVPR method, reconstructs event edge images and employs NetVLAD for descriptor extraction. In contrast, our Spike-EVPR leverages an end-to-end spiking architecture to extract descriptors, achieving 20.07\% higher Recall@1 on Brisbane-Event-VPR while offering superior energy efficiency.

On ss1/ss2 sequences, both Sparse-Event-VPR and Ev-ReconNet slightly outperform our method. This can be attributed to the low motion and consistent illumination between query and database views.
Under such conditions, Sparse-Event-VPR can take advantage of sparse event pixels, and Ev-ReconNet can reconstruct high-quality edge images for matching.
Nevertheless, our method consistently provides strong global descriptors across diverse environments.

\textbf{Note.} Neither Ev-ReconNet nor Sparse-Event-VPR provides results on DDD20: the former lacks open-source code, while the latter omits key DDD20-specific implementation details despite being open-sourced. FE-Fusion-VPR\cite{hou2023fe} is excluded here, as it uses both frames and events.

\subsection{Ablation studies}
To thoroughly analyze the effectiveness of each module and its design strategy in our method, we conduct ablation studies on Brisbane-Event-VPR and DDD20 datasets.
\subsubsection{Impact of Different Spike Representations}
\begin{table}[htbp]
\centering
\caption{Impact of spike representations on Recall@1(\%).}
\setlength{\tabcolsep}{8pt}
\scalebox{1.2}{
\begin{tabular}{ccc|cc}
\toprule[.05cm]
{SNN} & {MCS-T} & {TSS-T} & Brisbane & DDD20 \\
\hline
 & \checkmark &  & 65.83 & 26.29 \\
 \checkmark & \checkmark &  & 68.28 & 28.15 \\
 & & \checkmark   & 67.42 & 27.51 \\
 \checkmark & & \checkmark & 70.59 & 29.87 \\ 
 \checkmark & \checkmark & \checkmark & \textbf{74.46} &\textbf{33.35} \\ 
\toprule[.05cm]
\end{tabular}}
\label{tab:representations}
\vspace{-10pt}
\end{table}
We investigate the effect of our proposed spike representations and their ANN counterparts on EVPR, as shown in Tab.~\ref{tab:representations}. 
We remove the SSD-Extractor and CDA-Module, retaining only the BSR-Encoder. For each spike representation, global average pooling and L2 normalization are applied to generate the global descriptor. 
For the ANN variant, all spiking neurons are replaced with ANN counterparts. When using both representations, we sum the two descriptors to form the final descriptor.
The results in Tab.~\ref{tab:representations} show that using a single spike representation leads to a notable performance drop on both datasets. Moreover, the SNN-based configurations consistently outperform their ANN counterparts, demonstrating the benefits of SNNs for EVPR.

\begin{table*}[t]
\centering
\caption{Energy efficiency comparison between Spike-EVPR and its corresponding ANN models.}
\setlength{\tabcolsep}{13pt}
\scalebox{1.2}{
\begin{tabular}{ccccccc}
\toprule[.05cm]
{Models} & {Architecture} & {Params(M)} & {T} & $\mathrm{O}_{AC}$(G)
 & $\mathrm{O}_{MAC}$(G)
 & {Energy(mJ)}\\ 
 \hline
\multirow{2}{*}{ANN} & ResNet18  & 5.63 & - & - & 4.38 & 20.148\\
& ResNet34  & 16.40 & - & - & 8.61 & 39.606\\
 \hline
\multirow{2}{*}{SNN} & Sew-ResNet18  & 5.63 & 4  & 0.47 & 1.03 & 5.161\\
 & Sew-ResNet34  & 16.40  & 4   & 0.87   & 1.03  & 5.521\\
\toprule[.05cm]
\end{tabular}}
\label{tab:efficiency}
\vspace{-5pt}
\end{table*}
\subsubsection{Impact of Main Modules}
\begin{table}[htbp]
\centering
\caption{Impact of main modules on Recall@1(\%).}
\setlength{\tabcolsep}{4pt}
\scalebox{1.2}{
\begin{tabular}{cccc|cc}
\toprule[.05cm]
{MCS-T} & {TSS-T} & {SSD} & {CDA} & Brisbane & DDD20 \\
\hline
 & &  & & 71.46 & 28.13 \\
 \checkmark & & \checkmark & \checkmark & 74.42 & 32.13 \\
 & \checkmark & \checkmark & \checkmark & 75.15 & 34.42 \\
 \checkmark & \checkmark & \checkmark & & 77.42 & 35.22 \\
 \checkmark & \checkmark & & \checkmark & 78.59 & 37.60 \\ 
 \checkmark & \checkmark & \checkmark & \checkmark & \textbf{80.77} &\textbf{39.41} \\ 
\toprule[.05cm]
\end{tabular}}
\label{tab:Modules}
\end{table}
Tab.~\ref{tab:Modules} presents a breakdown analysis of the main components in Spike-EVPR. Without any modules, the baseline achieves 71.46\% and 28.13\% Recall@1 on the Brisbane and DDD20 datasets, respectively. 
Adding either MCS-T or TSS-T individually leads to substantial improvements, while using both spike representations together significantly boosts performance, highlighting their complementarity. 
Introducing the SSD-Extractor further improves performance by splitting and reassembling two spike features to generate multiple sub-descriptors with distinct characteristics.
Finally, the CDA-Module further improves performance by learning to weigh and aggregate the shared and specific sub-descriptors into a refined global descriptor. 
This learned aggregation helps the model better exploit complementary feature information, resulting in stronger place recognition performance.

\subsubsection{Impact of CDA-Module}
\begin{table}[t]
\centering
\caption{Impact of CDA-Module on Recall@1(\%) for shared and specific descriptors aggregation strategies.}
\setlength{\tabcolsep}{9pt}
\scalebox{1.2}{
\begin{tabular}{ccc|cc}
\toprule[.05cm]
{Concat} & {ADD} & {CDA} & Brisbane & DDD20 \\ 
\midrule
\checkmark &  &  & 78.42 & 36.13 \\
& \checkmark &  & 77.15 & 26.42 \\
&  & \checkmark & \textbf{80.77} &\textbf{39.41} \\
\toprule[.05cm]
\end{tabular}}
\label{tab:aggregation}
\vspace{-3pt}
\end{table}
We conduct an ablation study to evaluate the impact of the CDA-Module, as shown in Tab.~\ref{tab:aggregation}. 
We compare three different aggregation strategies for the shared and specific descriptors generated by the SSD-Extractor. 
The first simply concatenates them along the final dimension without using the CDA-Module. 
The second replaces the flatten operation in the module with a basic addition. 
The third uses the full CDA-Module as proposed.
The improved Recall@1 results suggest that the CDA-Module effectively learns how to balance the contributions of the shared and specific descriptors. This enables the final descriptor to capture richer features and focus on more discriminative information, similar to attention mechanisms.

\subsubsection{Impact of Geographic Distance Thresholds}
\begin{table}[t]
\begin{center}
\caption{Performance evaluation of Spike-EVPR under different geographic distance thresholds $\phi$.}
\setlength{\tabcolsep}{2.5pt}

\scalebox{1.1}{
\begin{tabular}{c|cccc|cc}
\toprule[.05cm]
\multirow{2}{*}{\makecell{Threshold\\$\phi$}}
&\multicolumn{4}{c|}{{Brisbane}}
&\multicolumn{2}{c}{{DDD20}}\\ 
\cline{2-7}
&ss1/ss2  &ss1/sr   &ss1/mn   &ss1/dt   &**83/**48  &**81/**68    \\
\midrule
15m &75.28  &76.21  &79.06  &43.21  &44.94  &6.05\\
30m &86.09  &84.64  &86.40  &55.32  &50.79  &14.24\\
45m &87.05  &85.84  &87.69  &57.98  &53.40  &17.79\\
60m &87.44  &86.16  &88.30  &58.85  &55.85  &19.63\\
75m &87.78  &86.56  &88.98  &59.77  &58.12  &21.48\\
\toprule[.05cm]
\end{tabular}}
\label{tab:4}
\end{center}
\vspace{-3pt}
\end{table}

\begin{figure}[t]
\centering
\includegraphics[width=0.45\textwidth]{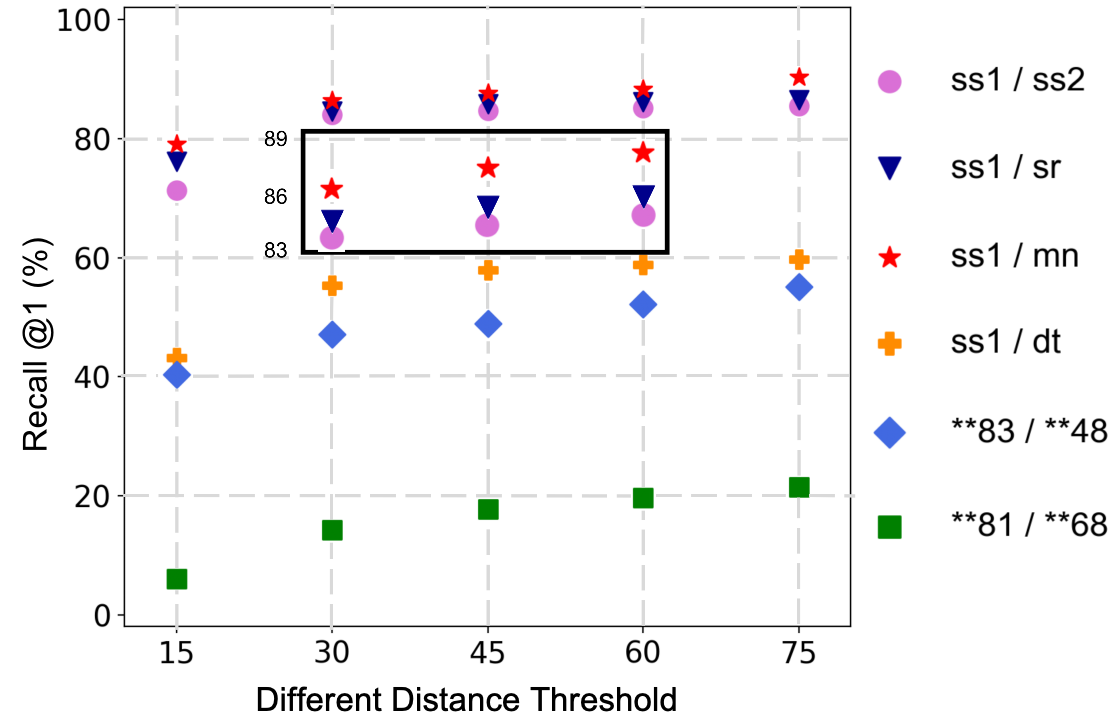}
\caption{Impact of different geographic distance thresholds.}
\label{fig:sensitivity}
\vspace{-3pt} 
\end{figure}
To evaluate the sensitivity of Spike-EVPR, we report Recall@1 under different geographic distance thresholds $\phi$ for true positives, as shown in Fig.~\ref{fig:sensitivity} and Tab.~\ref{tab:4}. As $\phi$ decreases, Recall@1 drops. Notably, the performance remains robust at 30m, with a slight decrease compared to the standard 75m setting, while a more substantial drop appears at 15m.
This is likely due to the coarse GPS sampling rate (intervals exceeding 2 seconds) and average vehicle speeds of 10-15m/s, causing adjacent GPS points to be spaced more than 15m apart, which can lead to inaccurate matching and reduced accuracy.
\begin{figure}[!t]
\centering
\begin{subfigure}[b]{0.23\textwidth}
\includegraphics[width=\textwidth]{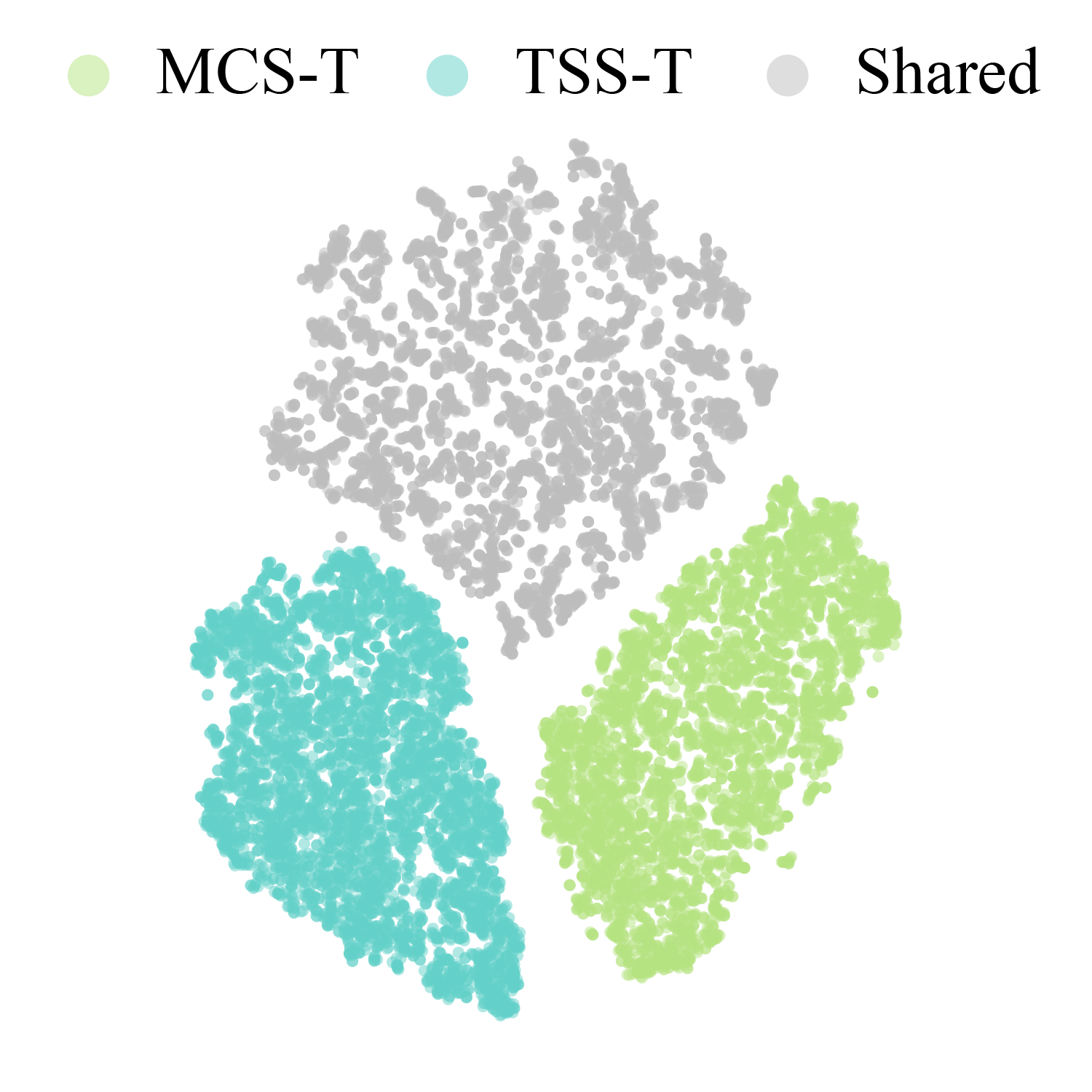}
\caption{Brisbane (ss1/mn)}
\end{subfigure}
\begin{subfigure}[b]{0.23\textwidth}
\includegraphics[width=\textwidth]{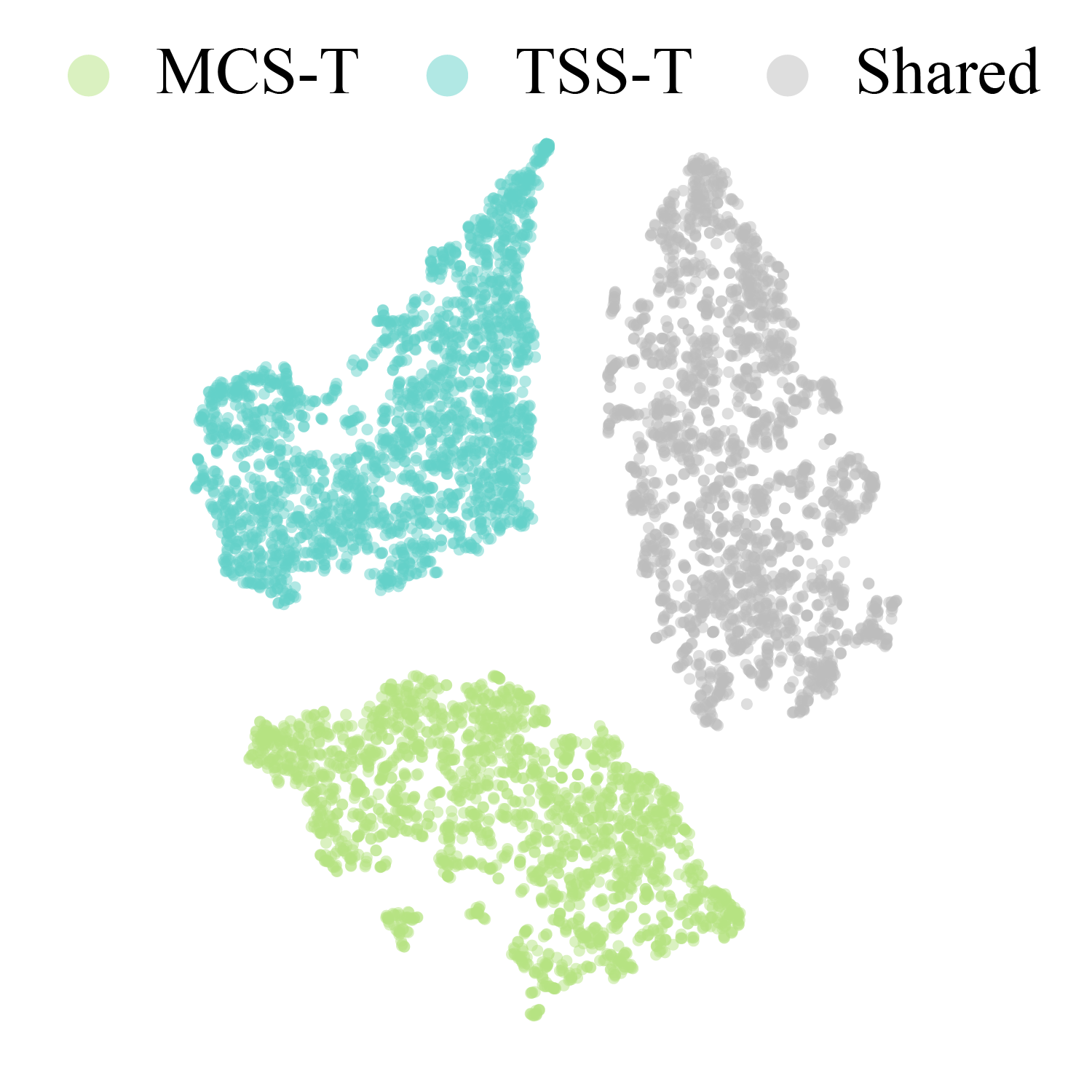}
\caption{DDD20 (**83/**48)}
\end{subfigure}
\caption{The t-SNE visualization of shared and specific features learned by the SSD-Extractor. `MCS-T' and `TSS-T' denote the specific features extracted from the MCS-Tensor and TSS-Tensor branches, respectively.}
\label{fig:tsne}
\end{figure}

\subsection{Feature Visualization}
To demonstrate the effectiveness of the SSD-Extractor, we visualize the learned shared and specific features using t-SNE~\cite{van2008visualizing}, as shown in Fig.~\ref{fig:tsne}. The shared features are compactly clustered, while the specific features extracted from MCS-Tensor and TSS-Tensor form separated distributions. This suggests that the two branches indeed learn complementary and non-redundant representations. 
\subsection{Energy Consumption Analysis}
We compute energy consumption based on operation counts under the 45nm CMOS technology~\cite{horowitz20141}, where each MAC and AC operation consumes 4.6pJ and 0.9pJ, respectively. In ANNs, each operation involves multiply-accumulate (MAC) operations, while in SNNs, neurons only perform accumulation (AC) operations when they spike, offering significant energy advantages. 
As shown in Tab.~\ref{tab:efficiency}, our Spike-EVPR built on Sew-ResNet18 and Sew-ResNet34~\cite{fang2021deep} consumes only 25.6\% and 13.9\% of the energy compared to ANN-based models using ResNet18 and ResNet34 backbones. 
Notably, MACs in our model are only involved during event-to-spike encoding, while the remaining network operates solely with ACs. 
This architecture ensures that increasing network depth does not incur higher computational cost, thus enabling both scalability and energy efficiency. These results demonstrate that our proposed Spike-EVPR achieves high energy efficiency.

\section{Conclusion}
In this paper, we address the challenges of applying SNNs to event-based VPR, focusing on spike representation and network architecture design. We propose Spike-EVPR, a directly trained SNN framework tailored for EVPR, which integrates spike-compatible representations and a deep spiking residual network with cross-representation aggregation to learn robust global descriptors.
Extensive experiments demonstrate that Spike-EVPR outperforms both ANN-based and SNN-based EVPR methods while reducing energy consumption. 
Future work will explore lightweight optimization and deployment on neuromorphic chips~\cite{jiang2023neuro,yu2023brain,jiang2025fully}.

\bibliographystyle{IEEEtran}
\bibliography{refer}

\end{document}